\documentclass{article}

\usepackage{arxiv}

\usepackage[utf8]{inputenc}
\usepackage[T1]{fontenc}
\usepackage{hyperref}
\usepackage{url}
\usepackage{booktabs}
\usepackage{amsfonts}
\usepackage{nicefrac}
\usepackage{microtype}
\usepackage{graphicx}
\usepackage{natbib}
\usepackage{doi}
\usepackage{tikz}
\usetikzlibrary{arrows.meta,positioning,fit,shapes.geometric}
\usepackage{float}
\usepackage{algorithm}
\usepackage{algpseudocode}
\usepackage{amsmath, amssymb, amsthm}
 \usepackage{physics}
\usepackage{algorithm}
\newcommand{\vect}[1]{\mathbf{#1}}
\newcommand{\mat}[1]{\mathbf{#1}}
\usepackage{romannum}
\usepackage{subcaption}

\title{X-VMamba: Explainable Vision Mamba}

\author{
  Mohamed Mabrok \\
  Department of Mathematics\\
  Qatar University\\
  Doha, Qatar \\
  \texttt{m.a.mabrok@qu.edu.qa} \\
  \And
  Yalda Zafari \\
  Department of Mathematics\\
  Qatar University\\
  Doha, Qatar \\
  \texttt{yaldazafari5@gmail.com} \\
}

\begin{document}
\maketitle

\begin{abstract}
State Space Models (SSMs), particularly the Mamba architecture, have recently emerged as powerful alternatives to Transformers for sequence modeling, offering linear computational complexity while achieving competitive performance. Yet, despite their effectiveness, understanding how these Vision SSMs process spatial information remains challenging due to the lack of transparent, attention-like mechanisms. To address this gap, we introduce a controllability-based interpretability framework that quantifies how different parts of the input sequence (tokens or patches) influence the internal state dynamics of SSMs. We propose two complementary formulations: a Jacobian-based method applicable to any SSM architecture that measures influence through the full chain of state propagation, and a Gramian-based approach for diagonal SSMs that achieves superior speed through closed-form analytical solutions. Both methods operate in a single forward pass with linear complexity, requiring no architectural modifications or hyperparameter tuning. We validate our framework through experiments on three diverse medical imaging modalities, demonstrating that SSMs naturally implement hierarchical feature refinement from diffuse low-level textures in early layers to focused, clinically meaningful patterns in deeper layers. Our analysis reveals domain-specific controllability signatures aligned with diagnostic criteria, progressive spatial selectivity across the network hierarchy, and the substantial influence of scanning strategies on attention patterns. Beyond medical imaging, we articulate applications spanning computer vision, natural language processing, and cross-domain tasks. Our framework establishes controllability analysis as a unified, foundational interpretability paradigm for SSMs across all domains. Code and analysis tools will be made available upon publication.

\end{abstract}

\section{Introduction}
Convolutional Neural Networks (CNNs) dominated computer vision and became the default architecture for different tasks such as classification, segmentation, and object detection. Their success stemmed from their inductive biases, translation equivariance through weight sharing and local receptive fields that mirrored how biological visual systems process information hierarchically. CNNs efficiently capture spatial hierarchies by stacking convolutional layers and learn simple patterns such as edge in early layers and complex one in deeper ones. However, CNNs faced a significant limitation: their local receptive fields restricted their ability to capture global context and modeling long-range dependencies across an image \cite{li2022modeling}. 

The introduction of Vision Transformers (ViTs) \cite{dosovitskiy2020image} marked a paradigm shift, adapting the transformer architecture from natural language processing \cite{vaswani2017attention} to visual tasks by treating images as sequences of patches. Unlike CNNs, ViTs employed self-attention mechanisms that allowed every patch to attend to every other patch from the first layer, enabling immediate global context modeling without the gradual expansion of receptive fields. Despite their groundbreaking performance, Transformers come with a significant drawback: the self-attention mechanism has a computational and memory complexity that scales quadratically ($O(N^2)$) with the sequence length $N$~\cite{tay2022efficient}. This quadratic scaling makes it prohibitively expensive to apply Transformers to the very long sequences encountered in high-resolution imaging data, such as medical images.

This situation created a fundamental trade-off in deep learning architecture design. An ideal model should be able to capture long-range dependencies, process local information with high fidelity, and maintain computational efficiency. While CNNs excel at efficient local feature extraction, they struggle with global context. Conversely, Transformers master global context but at a steep computational cost that becomes untenable for long sequences. This tension has motivated an active search for alternative architectures that can achieve the best of both worlds.

Recently, State Space Models (SSMs) have emerged as a powerful third paradigm. With roots in classical control theory dating back to Kalman's seminal work~\cite{kalman1960contributions}, SSMs model sequences by mapping a 1D input signal to an output signal through an intermediate latent state governed by linear dynamical systems. Early deep learning adaptations of SSMs, such as the Structured State Space for Sequences (S4) model, demonstrated that these models could be formulated as either a recurrent system for inference or a large convolutional kernel for training, enabling them to model long dependencies with near-linear complexity~\cite{gu2022efficiently}. Subsequent variants, including S5~\cite{smith2023simplified} and Liquid S4~\cite{hasani2022liquid}, further improved stability and expressiveness, establishing SSMs as a viable alternative to Transformers across multiple domains.

Building upon this foundation, the Mamba architecture introduced a critical innovation: a selective scan mechanism~\cite{gu2023mamba}. Unlike previous SSMs that were time-invariant (i.e., their state transition matrices $\mathbf{A}$, input matrices $\mathbf{B}$, and output matrices $\mathbf{C}$ were fixed), Mamba allows these parameters, and crucially, the discretization step size $\Delta$—to be input-dependent. This enables the model to selectively focus on or ignore parts of the input sequence, effectively compressing relevant context in its hidden state while allowing irrelevant information to decay rapidly. 

This content-aware reasoning allows Mamba to scale linearly with sequence length while demonstrating performance competitive with, and often superior to, Transformers on a variety of benchmarks spanning language modeling~\cite{gu2023mamba}, time-series forecasting~\cite{wang2024mamba}, image analysis~\cite{heidari2024computation}, and even reinforcement learning~\cite{huang2024decision}. This unique blend of efficiency, powerful representation learning, and selectivity has positioned Mamba and the broader family of SSMs as a leading contender for the next generation of foundational models, with transformative potential across a vast array of applications.

Despite the growing adoption of SSMs in various areas, a critical gap remains: \textit{we lack principled methods to understand what these models have learned and how they process information}. While Transformers benefit from a rich ecosystem of interpretability tools (attention visualization, attention rollout, probing classifiers), SSMs present unique challenges due to their fundamentally different information flow. Unlike Transformers, which aggregate information through explicit pairwise attention weights that can be visualized, SSMs propagate information \textit{implicitly} through hidden state dynamics governed by differential equations. The question "which input tokens influence the output?" cannot be answered by inspecting attention matrices, because there are none. Instead, we must analyze how inputs \textit{control} the evolution of the latent state over time, a problem that lies at the intersection of deep learning and dynamical systems theory.

This interpretability gap is particularly acute in high-stakes domains such as medical image analysis, clinical decision support, and autonomous driving, where regulatory bodies increasingly demand mechanistic explanations for AI-assisted decisions. For instance, if an SSM-based diagnostic model flags a patient scan as high-risk, clinicians need to understand \textit{which regions of the image} drove that prediction and \textit{how strongly} those regions influenced the model's internal reasoning. Existing explainability methods, designed primarily for CNNs and Transformers, fall short when applied to SSMs for three fundamental reasons:

(1) Techniques such as Grad-CAM~\cite{selvaraju2017grad}, Integrated Gradients~\cite{sundararajan2017axiomatic}, and SmoothGrad~\cite{smilkov2017smoothgrad} compute attributions by measuring how the output changes with respect to input perturbations via backpropagation. However, these methods suffer from gradient noise, saturation artifacts, and sensitivity to random initialization~\cite{adebayo2018sanity}. For SSMs, which involve exponential discretization ($\bar{\mathbf{A}} = \exp(\Delta \mathbf{A})$) and complex recurrent dynamics, gradients can explode or vanish, producing explanations that vary dramatically across runs or hyperparameter settings. Moreover, gradient methods require multiple forward-backward passes, making them computationally expensive for long sequences, precisely the regime where SSMs excel.

(2) A large body of work has focused on interpreting Transformers by analyzing attention weights~\cite{abnar2020quantifying}. However, SSMs do not have attention mechanisms; they propagate information through state transitions, not pairwise token interactions. Recent attempts to visualize SSM behavior have relied on ad-hoc heuristics such as plotting hidden state activations or measuring output sensitivity to input masking~\cite{ali2024hidden}, but these approaches lack theoretical grounding and fail to quantify how much each input controls the model's decision-making process.

(3) Even when gradient or perturbation-based methods produce plausible saliency maps, they are fundamentally post-hoc: they attempt to reverse-engineer what the model has already computed, rather than measuring the computation itself. As Jain and Wallace~\cite{jain2019attention} demonstrated for attention weights, high saliency does not guarantee high influence, subsequent layers may nullify the contribution of highly salient features. For SSMs, where information flows through a chain of state transitions ($h_t = \mathbf{A}h_{t-1} + \mathbf{B}u_t$), we need a method that directly measures \textit{controllability}: the degree to which an input at time $t$ can steer the hidden state trajectory and, consequently, the final output.

This work introduces a fundamentally different approach to SSM interpretability, grounded in classical control theory. Rather than approximating influence through gradients or heuristically interpreting hidden states, we leverage the concept of controllability  and controllability Gramian, mathematically rigorous metric that has governed dynamical systems analysis for over 60 years~\cite{kalman1960contributions}, to directly quantify how much each input element controls the internal state dynamics of the SSM. The Gramian $\mathcal{W}_c$ measures the "reachable state space" from a given input: if an input can strongly influence the hidden state (high controllability), it will also strongly influence the output, provided that the state is observable (weighted by the output matrix $\mathbf{C}$). We evaluated our proposed approach on three distinct medical imaging datasets, one grayscale mammography, one color dermatology, and one cellular microscopy, to test its generalization across radically different visual characteristics. We applied our method to the MedMamba architecture~\cite{yue2024medmamba}, a state-of-the-art SSM-based vision model trained on these datasets. Beyond vision tasks, our controllability framework opens new research directions for SSMs across all domains, such as language models, time-series forecasting, and multi-modality models, and reinforcement learning. Our controllability framework offers four critical advantages that distinguish it from all prior explainability methods:

\begin{enumerate}
    \item The controllability is not a post-hoc approximation; it is an intrinsic property of the state-space dynamics, derived from first principles in control theory. It is faithful by construction: high controllability scores necessarily imply high influence on the output (modulo observability, which we explicitly account for). This eliminates the faithfulness concerns that plague attention-based and gradient-based methods.
    
    \item By computing controllability at every layer of a multi-layer SSM, we reveal how information is progressively refined through the model's temporal hierarchy. Early layers capture low-level patterns (edges, textures), while deeper layers focus on semantic features (objects, diagnostic indicators). This layer-wise analysis, impossible with attention rollout in Transformers due to residual connections and non-linearities, provides unprecedented visibility into the model's internal reasoning process.
    
    \item The method is architecture-agnostic for any diagonal SSM, requiring no task-specific tuning, no hyperparameters, and no architectural modifications. It applies equally to language modeling (token-level influence), time-series forecasting (temporal importance), computer vision (spatial saliency), audio processing (spectro-temporal analysis), and genomics (base-pair relevance). The mathematical foundation is domain-independent, only the interpretation of "influence" changes across applications.

    \item  Influence scores are computed in a single forward pass with $O(N)$ complexity per sequence element, where $N$ is the state dimension (typically 8-64). Unlike gradient methods that require backpropagation through potentially unstable recurrent dynamics, or perturbation methods that require $O(L)$ forward passes for sequence length $L$.
\end{enumerate}

\section{State Space Models (SSMs)}
\label{sec:ssm_formulation}

At its core, a State Space Model conceptualizes a dynamic system by mapping an input sequence, $\mathbf{u}(t)$, to an output sequence, $\mathbf{y}(t)$, through a latent hidden state, $\mathbf{x}(t) \in \mathbf{R}^{N}$. This formulation is particularly powerful for modeling systems where the output depends not only on the current input but also on the history of all prior inputs, which is encapsulated within the state. The foundational SSM is described by a pair of linear ordinary differential equations (ODEs):

\begin{align}
\dot{\mathbf{x}}(t) &= \mathbf{A} \mathbf{x}(t) + \mathbf{B} \mathbf{u}(t) \label{eq:state_continuous} \\
\mathbf{y}(t) &= \mathbf{C} \mathbf{x}(t) + \mathbf{D} \mathbf{u}(t) \label{eq:output_continuous}
\end{align}

Here, $\mathbf{A} \in \mathbf{R}^{N \times N}$ is the state matrix that governs the evolution of the internal state, $\mathbf{B} \in \mathbf{R}^{N \times 1}$ is the input matrix that determines how the input influences the state, $\mathbf{C} \in \mathbf{R}^{1 \times N}$ is the output matrix that maps the state to the final output, and $\mathbf{D} \in \mathbf{R}^{1 \times 1}$ is the feed-forward matrix providing a direct link from input to output. For simplicity in many sequence modeling tasks, the feed-forward term $\mathbf{D}$ is often omitted.

While the continuous-time representation is mathematically elegant, deep learning models operate on discrete data. Therefore, the continuous parameters must be converted into discrete counterparts. This process, known as discretization, is typically performed using a timescale parameter, $\Delta$, which represents the sampling step. A standard technique for this conversion is the zero-order hold (ZOH), which yields the discrete-time matrices $\overline{\mathbf{A}}$ and $\overline{\mathbf{B}}$:

\begin{align}
\overline{\mathbf{A}} &= \exp(\Delta \mathbf{A}) \\
\overline{\mathbf{B}} &= (\Delta \mathbf{A})^{-1}(\exp(\Delta \mathbf{A}) - \mathbf{I}) \cdot \Delta \mathbf{B}
\end{align}

With these discrete parameters, the SSM can be expressed in two primary forms: a recurrent representation and a convolutional representation.

\textbf{Recurrent Form:} The discretized SSM can be computed sequentially, step-by-step, much like an RNN. The state at time step $t$ is calculated based on the previous state and the current input:
\begin{align}\label{eq:reqnt}
\mathbf{x}_t &= \overline{\mathbf{A}} \mathbf{x}_{t-1} + \overline{\mathbf{B}} \mathbf{u}_t \\
\mathbf{y}_t &= \mathbf{C} \mathbf{u}_t
\end{align}
This recurrent formulation is efficient during inference (auto-regressive generation) as it only requires the previous state. However, its sequential nature makes it inherently slow for training on parallel hardware like GPUs.

\textbf{Convolutional Form:} To overcome the training bottleneck of the recurrent form, the model can be unrolled. By recursively substituting the equation for $\mathbf{h}_t$, we can express the output $\mathbf{y}_t$ as a function of all previous inputs:
\begin{equation}
\mathbf{y}_t = \sum_{i=0}^{t} \mathbf{C} \overline{\mathbf{A}}^{t-i} \overline{\mathbf{B}} \mathbf{x}_i
\end{equation}
This formulation reveals that the SSM is equivalent to a discrete convolution. The entire output sequence $\mathbf{y}$ can be computed in parallel by convolving the input sequence $\mathbf{x}$ with a structured convolutional kernel $\overline{\mathbf{K}}$:
\begin{equation}
\mathbf{y} = \mathbf{x} * \overline{\mathbf{K}}
\end{equation}
where the kernel $\overline{\mathbf{K}}$ of length $L$ is defined as:
\begin{equation}
\overline{\mathbf{K}} = (\mathbf{C} \overline{\mathbf{B}}, \mathbf{C} \overline{\mathbf{A}} \overline{\mathbf{B}}, \dots, \mathbf{C} \overline{\mathbf{A}}^{L-1} \overline{\mathbf{B}})
\end{equation}
This convolutional representation allows for highly parallelized and efficient training using standard deep learning frameworks and Fast Fourier Transforms (FFTs). However, a key limitation of this traditional SSM is that its parameters ($\mathbf{A}, \mathbf{B}, \mathbf{C}$) are fixed and independent of the input data, making it a Linear Time-Invariant (LTI) system. This property restricts its capacity to model complex, content-dependent dynamics, a challenge directly addressed by the selective mechanisms in Mamba.

The Vision Mamba (VMamba) architecture proposes visual state space components specifically designed to handle image-based inputs \cite{liu2024vmamba}. The fundamental innovation enabling this adaptation is the 2D Selective Scan (SS2D) mechanism, which employs four different scanning patterns across the divided image patches. These varied traversal approaches enable the architecture to preserve complex spatial relationships that typically deteriorate when converting images into one-dimensional sequential representations. Through maintaining the inherent two-dimensional organization and integrating directional preferences, SS2D strengthens the model's ability to represent spatial information effectively.

\section{Explainability Foundation for SSMs}
\label{sec:method}

Controllability, in control theory, refers to the ability to steer a system's state to any desired value. We re-purpose this idea to quantify the "steering power" or influence of each input token or patch, and we introduce a measure named \textit{influence score}. The influence score is defined as how much effect a given token or image patch has on the final output. We define our problem formulation as follows: 

Given an SSM model processing an input sequence $\mathbf{X} \in \mathbf{R}^{H \times W \times C}$, we seek to compute an influence score $s_{i,j} \in \mathbf{R}^+$ for each spatial token or location $(i,j)$ that quantifies: (1) how much the input at location $(i,j)$ controls the hidden state evolution, (2) how observable this control is in the output space, and (3) the cumulative effect across all state dimensions and scanning directions. We propose three different formulations of the influence score. 
 
\subsection{Initial Formulation: A Direct Application of Controllability Theory}
The most direct application of controllability theory to interpret a SSM involves quantifying the influence of an input, $\vec{u}_k$, on the system's terminal state, $\vec{x}_L$. In classical control systems, the ability to drive the final state to a desired configuration is the very definition of controllability. Therefore, a logical first step is to formulate an influence score based on the magnitude of the transformation mapping an input patch to this final state vector. This approach is mathematically attractive as it provides a direct measure of an input's power to steer the model's ultimate internal representation. To derive this relationship, we unroll the state update recurrence:
\begin{equation*}
    \vec{x}_{k+1} = \bar{\mathbf{A}}_k \vec{x}_k + \bar{\mathbf{B}}_k \vec{u}_k
\end{equation*}
The final state $\vec{x}_L$ can be expressed as a function of an initial state $\vec{x}_0$ and the sequence of all inputs $\vec{u}_1, \dots, \vec{u}_L$. The specific component of $\vec{x}_L$ that is linearly dependent on a past input $\vec{u}_k$ is isolated by tracing its path through the sequence of state transitions:
\begin{equation}
    \text{path}(\vec{u}_k \to \vec{x}_L) = \left( \prod_{j=k+1}^{L-1} \bar{\mathbf{A}}_j \right) \bar{\mathbf{B}}_k \vec{u}_k
    \label{eq:path_final_state_revised}
\end{equation}
The matrix product $\left( \prod_{j=k+1}^{L-1} \bar{\mathbf{A}}_j \right) \bar{\mathbf{B}}_k$ is the linear operator that transforms the input vector $\vec{u}_k$ into its contribution to the final state $\vec{x}_L$. Consequently, a natural definition for the influence score is the norm of this operator, which quantifies the amplification or attenuation of the input along this dynamical path.

While this formulation is theoretically sound from a classical control perspective, its practical application to deep, stable SSMs reveals a critical emergent property. The state matrices, $\mathbf{A}$, in architectures like Mamba are initialized and structured to ensure stability, which implies that the eigenvalues of their discretized counterparts, $\bar{\mathbf{A}}_k$, generally have magnitudes less than or equal to one. When computing the long product of these matrices as required by Eq. \ref{eq:path_final_state_revised}, the norm of the product decays exponentially with the length of the chain ($L-1-k$). This phenomenon, which we term \textit{vanishing influence}, is analogous to the vanishing gradient problem in deep recurrent neural networks. It leads to a severe recency bias in the calculated scores:
\begin{itemize}
    \item \textbf{Early-sequence inputs ($k \ll L$):} The transformation matrix for these inputs involves a long product of stable matrices, causing its norm to diminish toward zero. As a result, the model appears insensitive to features at the beginning of the sequence.
    \item \textbf{Late-sequence inputs ($k \approx L$):} The matrix product is short or empty (an identity matrix for $k=L-1$), resulting in a significantly larger norm. The model thus appears hypersensitive to features at the end of the sequence.
\end{itemize}
Empirical analysis confirms this hypothesis, yielding heatmaps where influence scores are overwhelmingly concentrated in the corners of the image, corresponding to the start and end points of the various 2D scanning paths. This outcome is an artifact of the analysis method itself rather than a reflection of content-based reasoning by the model. This critical finding motivates the development of a more sophisticated metric that is aligned with the model's specific architectural properties, particularly its use of global information aggregation for classification.

\subsection{ Influence on the Aggregated Output (Jacobian Method)}

The fundamental flaw in measuring influence on the final hidden state, $\vec{x}_L$, is that it is not the direct input to the classifier. VSSMs almost universally employ a Global Average Pooling (GAP) layer after the final SSM layer. The GAP layer computes the mean of the output vectors, $\vec{y}_k$, across all $L$ sequence steps (patches). This produces a single, fixed-size feature vector, $\bar{\vec{y}}$, which is then fed into the final linear classification head.
\begin{equation}
    \bar{\vec{y}} = \frac{1}{L} \sum_{j=1}^{L} \vec{y}_j = \frac{1}{L} \sum_{j=1}^{L} (\bar{\mathbf{C}}_j \vec{x}_j + \bar{\mathbf{D}}_j \vec{u}_j)
    \label{eq:gap}
\end{equation}
Therefore, to create a meaningful interpretability metric, we must measure the influence of an input patch $\vec{u}_k$ on this aggregated vector $\bar{\vec{y}}$. An input $\vec{u}_k$ influences the aggregated output $\bar{\vec{y}}$ by contributing to its own output term, $\vec{y}_k$, and to all subsequent output terms, $\vec{y}_j$ where $j > k$. The total influence is the sum of the magnitudes of these individual contributions.

Let's first define the state propagation. The state $\vec{x}_j$ at any step $j$ is a function of a prior state $\vec{x}_k$ ($k < j$) and the intervening inputs:
\begin{equation}
    \vec{x}_j = \left( \prod_{i=k}^{j-1} \bar{\mathbf{A}}_i \right) \vec{x}_k + \sum_{m=k}^{j-1} \left( \prod_{i=m+1}^{j-1} \bar{\mathbf{A}}_i \right) \bar{\mathbf{B}}_m \vec{u}_m
\end{equation}
The partial derivative $\frac{\partial \vec{x}_j}{\partial \vec{u}_k}$ gives us the linear transformation that maps a change in input $\vec{u}_k$ to a change in a future state $\vec{x}_j$:
\begin{equation}
    \frac{\partial \vec{x}_j}{\partial \vec{u}_k} = \left( \prod_{i=k+1}^{j-1} \bar{\mathbf{A}}_i \right) \bar{\mathbf{B}}_k \quad \text{for } j > k
    \label{eq:state_jacobian}
\end{equation}
Similarly, the influence of $\vec{u}_k$ on an output $\vec{y}_j$ is found by applying the chain rule:
\begin{equation}
    \frac{\partial \vec{y}_j}{\partial \vec{u}_k} = \frac{\partial \vec{y}_j}{\partial \vec{x}_j} \frac{\partial \vec{x}_j}{\partial \vec{u}_k} = \bar{\mathbf{C}}_j \left[ \left( \prod_{i=k+1}^{j-1} \bar{\mathbf{A}}_i \right) \bar{\mathbf{B}}_k \right]
    \label{eq:output_jacobian}
\end{equation}
For the special case where $j=k$, the influence is direct (it does not pass through $\mathbf{A}$):
\begin{equation}
    \frac{\partial \vec{y}_k}{\partial \vec{u}_k} = \bar{\mathbf{C}}_k \frac{\partial \vec{x}_k}{\partial \vec{u}_k} + \bar{\mathbf{D}}_k = \bar{\mathbf{C}}_k \bar{\mathbf{B}}_k + \bar{\mathbf{D}}_k
\end{equation}
Assuming the feedthrough matrix $\mathbf{D}$ is negligible or zero (common in these architectures), this simplifies to $\bar{\mathbf{C}}_k \bar{\mathbf{B}}_k$.

The total influence of $\vec{u}_k$ is the sum of the norms (we use the Frobenius norm, $\|\cdot\|_F$) of these Jacobian matrices for all outputs from $k$ to $L$:
\begin{equation}
    \text{InfluenceScore}(k) = \sum_{j=k}^{L} \left\| \frac{\partial \vec{y}_j}{\partial \vec{u}_k} \right\|_F
\end{equation}
Expanding this gives our final, detailed formula:
\begin{equation}
    \text{InfluenceScore}(k) = \underbrace{\| \bar{\mathbf{C}}_k \bar{\mathbf{B}}_k \|_F}_{\text{Term 1: Direct Influence}} + \underbrace{\sum_{j=k+1}^{L} \left\| \bar{\mathbf{C}}_j \left( \prod_{i=k+1}^{j-1} \bar{\mathbf{A}}_i \right) \bar{\mathbf{B}}_k \right\|_F}_{\text{Term 2: Propagated Influence}}
    \label{eq:final_detailed}
\end{equation}

The \textit{Direct Influence} ($\|\bar{\mathbf{C}}_k \bar{\mathbf{B}}_k\|_F$) measures the immediate impact of the input $\vec{u}_k$ on its own corresponding output $\vec{y}_k$.
\begin{itemize}
    \item $\bar{\mathbf{B}}_k$: This matrix transforms the input patch $\vec{u}_k$ into a change in the hidden state $\vec{x}_k$. A large norm for $\bar{\mathbf{B}}_k$ means the input is salient and can strongly perturb the state.
    \item $\bar{\mathbf{C}}_k$: This matrix reads out the hidden state $\vec{x}_k$ to produce the output $\vec{y}_k$. A large norm for $\bar{\mathbf{C}}_k$ means the current state is considered important for the output.
\end{itemize}
The product $\bar{\mathbf{C}}_k \bar{\mathbf{B}}_k$ represents the full, instantaneous input-output path at step $k$. This term ensures that every patch has a baseline influence score, preventing the "vanishing influence" problem for early patches.

The \textit{Propagated Influence} is a sum that captures the long-term impact of $\vec{u}_k$ on all subsequent outputs. For each future step $j > k$, the matrix product within the norm represents the dynamical path from input $\vec{u}_k$ to output $\vec{y}_j$.
\begin{itemize}
    \item $\left( \prod_{i=k+1}^{j-1} \bar{\mathbf{A}}_i \right)$: This is the state transition matrix that describes how the initial state perturbation caused by $\vec{u}_k$ evolves and propagates through the system's memory over time.
    \item The entire term $\bar{\mathbf{C}}_j (\dots) \bar{\mathbf{B}}_k$ is the linear operator mapping the input at time $k$ to the output at a future time $j$.
\end{itemize}
This sum  solves the recency bias. An early patch ($k \ll L$) has its influence on any single distant output $\vec{y}_L$ diminished by decay, but it has the opportunity to influence a large number of subsequent outputs, and the sum of these contributions can be substantial. The following algorithm shows how to compute the influence score.

\begin{algorithm}
\caption{Aggregated Controllability Analysis for Vision SSMs}
\begin{algorithmic}[1]
\State \textbf{Input:} Image feature map $\vect{X} \in \mathbb{R}^{H \times W \times C_{in}}$
\State \textbf{Output:} Influence Score Map $\mat{S} \in \mathbb{R}^{H \times W}$
\State Initialize $\mat{S}_{total} = \vect{0}^{H \times W}$
\For{each scan direction $d \in \{\text{fwd, bwd, transp-fwd, transp-bwd}\}$}
    \State Prepare 1D input sequence $\vect{U}^{(d)} = \{\vect{u}_1, \dots, \vect{u}_L\}$ by scanning $\vect{X}$
    \State Compute input-dependent SSM matrices for the sequence: $\{\bar{\mat{A}}_k, \bar{\mat{B}}_k, \bar{\mat{C}}_k\}_{k=1}^L$
    \State Initialize score map for this direction, $\mat{S}^{(d)} = \vect{0}^{L}$
    \State Initialize future influence propagator, $\vect{P} = \vect{0}$
    \For{$k = L \to 1$ (iterating backwards)}
        \State $\text{direct\_influence} = \| \bar{\mat{C}}_k \bar{\mat{B}}_k \|_F$
        \State $\text{propagated\_influence} = \| \vect{P} \cdot \bar{\mat{B}}_k \|_F$
        \State $S^{(d)}_k = \text{direct\_influence} + \text{propagated\_influence}$
        \State $\vect{P} \leftarrow \bar{\mat{C}}_k + \bar{\mat{A}}_k \cdot \vect{P}$ \Comment{Update propagator for next iteration}
    \EndFor
    \State Re-orient $\mat{S}^{(d)}$ from 1D sequence to 2D map according to scan direction $d$
    \State $\mat{S}_{total} \leftarrow \mat{S}_{total} + \mat{S}^{(d)}$
\EndFor
\State $\mat{S} \leftarrow \frac{1}{4} \mat{S}_{total}$
\State \textbf{return} $\mat{S}$
\end{algorithmic}
\end{algorithm}

\subsection{Gramian-Based Influence Score (Gramian Method )}

From classical control theory, the controllability Gramian over horizon $[0, T]$ can be computed:
\begin{equation}
    \mathcal{W}_c^{(T)}=\sum_{t=0}^{T-1} \mathbf{A}^t \mathbf{B} \mathbf{B}^T\left(\mathbf{A}^T\right)^t,
\end{equation}
where,  $\mathcal{W}_c^{(T)} \in \mathbb{R}^{N \times N}$ (symmetric positive semi-definite). This physically  can be understood as that the matrix  $\mathcal{W}_c$ quantifies the reachable state energy. In other words, ff $\mathcal{W}_c$ is full rank, any state $\mathbf{x}^* \in \mathbb{R}^N$ can be reached and  the eigenvalues of $\mathcal{W}_c$ indicate how much energy is needed to reach states in different directions.

For stable systems $\left(\left|\lambda_i(\mathbf{A})\right|<1\right)$, we can take $T \rightarrow \infty$ :

\begin{equation}
\mathcal{W}_c=\lim _{T \rightarrow \infty} \mathcal{W}_c^{(T)}=\sum_{t=0}^{\infty} \mathbf{A}^t \mathbf{B} \mathbf{B}^T\left(\mathbf{A}^T\right)^t.
\end{equation}
Which can be obtained by solving This is the continuous Lyapunov equation:
\begin{equation}
\mathbf{A} \mathcal{W}_c \mathbf{A}^T-\mathcal{W}_c+\mathbf{B} \mathbf{B}^T=\mathbf{0}
\end{equation}

Practically speaking, in many SSM implementation, the matrix $A$  is assumed to be diagonal. In this case the Gramian matrix becomes:
\begin{equation}
\mathcal{W}_c=\sum_{t=0}^{\infty} \mathbf{A}^{2 t} \mathbf{B} \mathbf{B}^T
\end{equation}
We can rewrite this as per-state, element-wise, and per location $l$ as follows:

\begin{equation}
W_{c, i}^{(k)}(l)=\frac{\sum_{m=1}^M\left[B^{(k)}(l)\right]_{i m}^2}{1-\left[a^{(k)}(l)\right]_i^2+\epsilon}=\frac{\left\|\mathbf{B}_{i,:}^{(k)}(l)\right\|^2}{1-a_i^{(k) 2}(l)+\epsilon}
\end{equation}

Now, we  weight the above by the observability, we get:

\begin{equation}
\mathcal{I}_i^{(k)}(l)=\left[C^{(k)}(l)\right]_i^2 \cdot W_{c, i}^{(k)}(l)
\end{equation}

Finally, we sum over state dimensions and average over inner channels:
\begin{equation}
    s^{(k)}(l) = \frac{1}{D} \sum_{d=1}^{D} \sum_{n=1}^{N} \mathcal{I}_n^{(k)}(l, d)
\end{equation}
where $D$ is the inner dimension (expansion factor $\times$ model dimension). The final score $s_{i,j}$ has a clear interpretation:

\begin{itemize}
    \item \textbf{High score}: The patch at $(i,j)$ has strong control over state dynamics ($B$ is large), the dynamics are persistent ($\bar{\mathbf{A}}$ is close to 1), and these states are highly observable in the output ($C$ is large).
    
    \item \textbf{Low score}: The patch has weak influence (small $B$), rapid decay (small $\bar{\mathbf{A}}$), or low observability (small $C$).
\end{itemize}

This differs from attention mechanisms which measure pairwise token interactions. Our controllability scores capture \textit{state-space influence}, how patches control the temporal evolution of hidden representations.

The primary theoretical distinction between the two analysis methods lies in their fundamental assumptions regarding the state transition matrix, $A_{\text{bar}}$. The Gramian-based method is intrinsically dependent on a diagonal $A_{\text{bar}}$. This assumption is what permits the calculation of the controllability Gramian $W_c$ via the closed-form, element-wise solution $W_c = \frac{B^2}{1 - A_{\text{bar}}^2}$. This simplification is computationally powerful but cannot be applied to a system with a non-diagonal state matrix, as that would necessitate solving a full, complex Lyapunov equation.

In sharp contrast, the Jacobian-based method is a theoretically general approach that does not fundamentally require $A_{\text{bar}}$ to be diagonal. This method traces the exact influence of an input $u_k$ on a future output $y_j$ by computing the propagated effect, $\frac{\partial y_j}{\partial u_k} \propto C_j \left(\prod_{i=k+1}^{j-1} A_i\right) B_k$. For a non-diagonal system, this state propagation ($\prod A_i$) would be correctly computed using full matrix multiplication, accurately modeling the complex interactions between all state dimensions. 

\section{Experimental Evaluation on Medical Image Classification}
\label{sec:experiments}

\subsection{Datasets and Model}
We employed the MedMamba architecture~\cite{yue2024medmamba}, a state-of-the-art model for medical classification that demonstrated superior performance compared to several benchmarks on different datasets. We trained the model on three different datasets, each with distinct features, to evaluate the proposed approach’s generalization. The datasets utilized are as follows:

\textbf{Mammography:} The CMMD dataset was used, which contains diagnostic mammography images~\cite{cui2021chinese}. Mammography images are grayscale and high-resolution, where subtle changes define the diagnostic features. Controllability-based analysis should enable us to assess the model’s ability to identify and capture the regions that a radiologist would flag. Additionally, detecting these subtle changes and abnormalities may be possible by comparing the characteristics of specific regions with those of normal regions. The model also needs to identify the relevant normal tissues for this comparison.

\textbf{Dermatology:} We used the DermaMNIST dataset from MedMNIST for skin lesion classification \cite{yang2021medmnist, yang2023medmnist}. It contains RGB images with rich texture and color information. The ABCDE criteria (Asymmetry, Border irregularity, Color variation, Diameter, and Evolution) are the main factors for assessment and diagnosis.

\textbf{Hematology:} BloodMNIST from MedMNIST was the third dataset used in this study, containing blood cell microscopic images \cite{yang2021medmnist, yang2023medmnist}. In these images, there are dense and repetitive structures (cells) with subtle morphological differences. The model needs to selectively attend to abnormal cells based on their structure and suppress redundant information from normal cells.

\subsection{Results}
We evaluated to Jacobian-based approach as it is more broad and doesn't contain the assumption of Gramian-based model, the diagonal $A$ matrix. We evaluated the controllability pattern evolution, influence score distribution, and more detailed analysis of influence score for different image scanning scheme in Mamba models.

\begin{figure}
    \centering
    \includegraphics[width=1\linewidth]{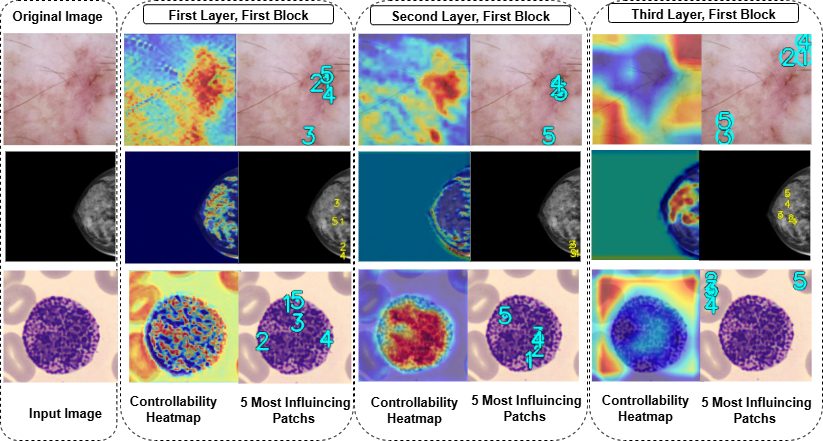}
    \caption{Heatmap of classification SSM model using several datasets in different architectural layers.}
    \label{fig:multi_dataset_controllability}
\end{figure}

Figure~\ref{fig:multi_dataset_controllability} illustrates the evolution of controllability patterns across the first three hierarchical layers of the VSSM-based modules for representative samples from each modality. A consistent pattern emerges across all three imaging modalities: controllability maps become progressively more focused and semantically meaningful with increasing depth, confirming the expected behavior of deep hierarchical models. In the first layer (leftmost panels), controllability scores remain relatively diffuse, capturing low-level textural features such as edges, intensity gradients, and local contrast variations. This observation aligns with the expectation that early SSM layers process fine-grained spatial information without strong semantic differentiation.

As the network deepens, the model begins to exhibit spatial selectivity, with higher controllability scores concentrating on diagnostically relevant regions. In dermatology images, this transition is evident as features shift from capturing diffuse skin textures to precisely localizing lesion boundaries and pigmentation patterns. By the third layer, sharp demarcation appears around the lesion perimeter, indicating that the model has learned to prioritize morphological boundaries critical for melanoma classification. Similarly, in mammography images, the focus evolves from diffuse fibroglandular tissue to concentrated attention on fat-tissue interfaces and ultimately on fibroglandular regions surrounding mass abnormalities. The heatmaps increasingly align with clinically relevant areas where radiologists focus during diagnostic assessment. In blood cell images, this same hierarchical refinement is observed: early layers distribute attention uniformly across multiple cells, while deeper layers selectively enhance individual cells or cell clusters based on diagnostic significance.

This progression demonstrates that Mamba's selective state-space mechanism naturally implements a coarse-to-fine attention refinement strategy, analogous to the receptive field hierarchy in convolutional networks but achieved through temporal dynamics rather than spatial convolution. Importantly, despite this universal hierarchical pattern, each imaging modality exhibits a distinct controllability signature that reflects domain-specific diagnostic features. In dermatological images, the controllability maps reveal strong weighting for color heterogeneity and border irregularity—key indicators for melanoma diagnosis. The model's emphasis on lesion boundaries in later layers further demonstrates its ability to extract relevant diagnostic features. In mammography, the heatmaps indicate that classifications are primarily driven by density distribution patterns, with the model initially detecting fibroglandular and fat tissue before shifting attention toward suspicious high-density areas. By layer 3, the most influential patches cluster around the abnormality, validating that the model has learned clinically meaningful patterns for mammography interpretation.

\begin{figure}
    \centering
    \includegraphics[width=1\linewidth]{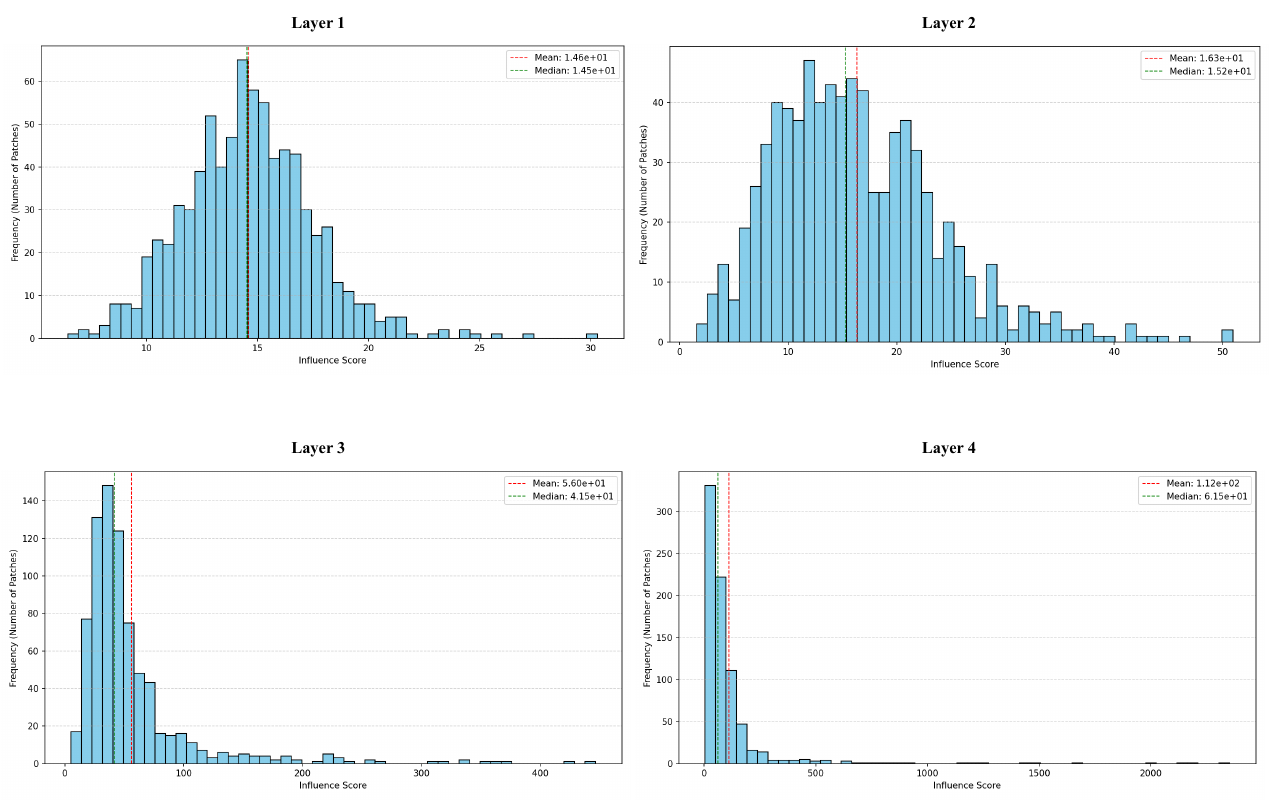}
    \caption{An illustration of influence score distribution across different layers of the model.}
    \label{fig:influence_score_distribution}
\end{figure}

Figure~\ref{fig:influence_score_distribution} illustrates the mean influence score distribution across different layers of the model. In the early layers, most patches contribute nearly equally to the model's output, reflecting the diffuse attention pattern observed in initial feature processing. As the network deepens, however, a clear selectivity emerges: the number of high-influence patches progressively decreases, while their maximum influence scores increase substantially. This trend indicates that deeper layers concentrate decision-making authority on a select subset of patches, with only a few regions ultimately dominating the model's classification process. The simultaneous reduction in the breadth of influence and amplification of peak influence scores demonstrates a natural hierarchical filtering mechanism, wherein the model progressively distills spatial information from broadly distributed low-level features to a sparse set of semantically significant regions that drive final predictions.

\begin{figure}
    \centering
    \includegraphics[width=1\linewidth]{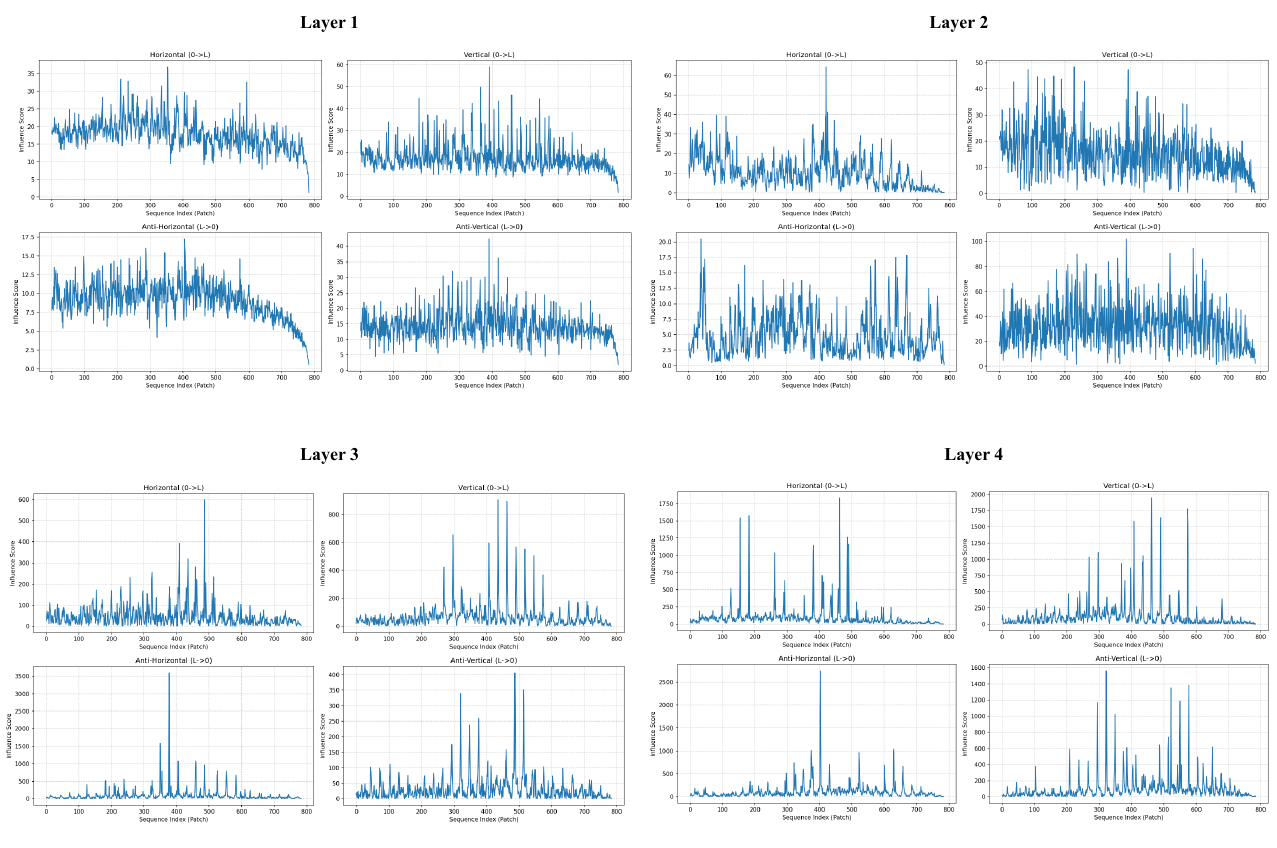}
    \caption{An illustration of influence score distribution across different image scanning schemes within different layers of the model.}
    \label{fig:influence_score_scanning}
\end{figure}

Figure~\ref{fig:influence_score_scanning} illustrates the influence impact of individual patches across different image scanning schemes employed in VMamba models throughout the network hierarchy. The results reveal that influence score distributions are substantially modulated by the choice of scanning strategy, indicating that directional sequencing fundamentally shapes the model's spatial attention patterns. Notably, across all scanning strategies, the characteristic progression from diffuse influence scores in early layers to concentrated, localized scores in deeper layers remains evident, confirming that this hierarchical refinement is a robust property of the architecture. These findings enable more detailed analysis of how image scanning strategies influence model decision-making processes and provide insights for developing more efficient scanning approaches that optimize diagnostic feature extraction.

\section{Potential Applications}
\label{sec:more_app}
\subsection{Computer Vision Applications}
The controllability-based interpretability framework opens numerous avenues for advancing vision systems beyond medical imaging, spanning autonomous systems, multimedia analysis, and model security. Its unique ability to trace information flow through temporal dynamics positions it as a foundational tool for understanding modern vision architectures.

\subsubsection{Video Understanding and Temporal Reasoning}
In video analysis tasks including action recognition, event detection, and video captioning, temporal dependencies fundamentally shape model predictions. Unlike frame-level attention mechanisms that treat each frame independently and aggregate their contributions post-hoc, our state-space approach can be extended and directly measures how information propagates through the temporal sequence. The controllability framework can reveal  which frames are important, and how early visual cues cascade through the hidden state to influence final decisions. For instance, in action recognition, controllability analysis can identify the exact moment when an action transitions from preparation to execution, quantifying how strongly this pivotal frame controls subsequent state evolution and prediction confidence.

The temporal propagation metric, $\sum_{j=k+1}^{L} \left\| \bar{\mathbf{C}}_j \left( \prod_{i=k+1}^{j-1} \bar{\mathbf{A}}_i \right) \bar{\mathbf{B}}_k \right\|_F$, provides a natural measure of temporal influence decay, revealing whether models appropriately weight recent versus historical frames. This capability addresses critical questions in video understanding: Does the model exhibit excessive recency bias, over-emphasizing the final frames? Does it correctly identify and maintain influence from semantically critical moments that occurred seconds earlier? Such analysis could improve video summarization by identifying frames with maximum downstream influence, enhance temporal action localization by measuring state controllability at action boundaries, and optimize video compression by preserving high-controllability frames while aggressively compressing low-influence regions.

\subsubsection{Autonomous Driving and Safety-Critical Perception}
In autonomous driving systems, where SSM-based perception models process sequential streams of camera, LiDAR, and radar data, interpretability is legally mandated for deployment. Our controllability framework addresses regulatory requirements for explainable AI in safety-critical domains by providing mechanistic explanations grounded in dynamical systems theory rather than heuristic post-hoc approximations. When an autonomous vehicle makes a critical decision (emergency braking, lane change, or yielding to a pedestrian) controllability maps can identify which spatial regions (pedestrian location, adjacent vehicle trajectory, road boundary) and which temporal frames (the moment a pedestrian steps off the curb) most strongly controlled the hidden state evolution leading to that decision.

This transparency enables engineers to verify that models focus on causally relevant features rather than spurious correlations. Furthermore, by comparing controllability maps across different weather conditions, lighting scenarios, and traffic densities, developers can identify domain shift vulnerabilities where models inappropriately redistribute attention under distribution shifts. The scanning direction analysis in VMamba architectures becomes particularly salient for autonomous perception, as different directional sweeps may capture distinct spatial relationships in structured driving environments. For instance, forward-backward scans naturally align with road geometry and traffic flow direction, while transposed scans might better capture lateral relationships between vehicles in adjacent lanes. Controllability analysis can quantify these directional biases, informing the design of scan augmentation strategies or adaptive scanning policies that optimize spatial attention for driving scenarios.

\subsubsection{Adversarial Robustness and Model Security}
he security and robustness of deep learning models can be rigorously assessed by studying their behavior under adversarial attacks through the lens of controllability analysis. This application transforms interpretability from a diagnostic tool into a forensic instrument for understanding vulnerability mechanisms.  Let $\vec{x}$ be a benign image and $\vec{x}_{adv}$ be its adversarially perturbed version that causes a misclassification. By computing their respective controllability maps, $\mat{S}$ and $\mat{S}_{adv}$, and analyzing the difference map, $\Delta\mat{S} = \mat{S}_{adv} - \mat{S}$, one can identify the specific spatial regions and hierarchical layers where the attack exerts maximum influence on state dynamics. This analysis reveals the mechanism of adversarial exploitation with unprecedented precision. Traditional gradient-based saliency methods show where the attack perturbs pixels, but controllability analysis reveals how those perturbations propagate through the state-space dynamics to corrupt the hidden representation.

\subsubsection{Remote Sensing and Large-Scale Spatial Analysis}
In satellite and aerial imagery analysis, SSM-based models process multi-spectral, hyperspectral, or temporal sequences of Earth observation data for applications including land cover classification, change detection, disaster assessment, and agricultural monitoring. The controllability framework addresses unique interpretability challenges in this domain, where inputs span multiple spectral bands (visible, infrared, radar) and temporal observations (daily, seasonal time series). By computing band-specific controllability scores, we can identify which spectral channels most strongly control predictions for each task. This spectral influence analysis guides optimal sensor selection and data fusion strategies.

\subsection{Language Model Applications}

Although our experiments focus on vision tasks, the controllability framework directly extends to SSM-based language models. It provides a principled way to measure how tokens, passages, and reasoning steps influence state dynamics, enabling mechanistic insight into long-context handling, prompt design, reasoning, and grounding.

\subsubsection{Long-Context Understanding and Information Retrieval}
Long-context SSMs such as Mamba can process documents or conversations far beyond typical Transformer limits. Controllability scores provide a principled way to identify which tokens or passages meaningfully affect the model's state evolution, directly addressing phenomena like “lost in the middle.” By measuring both immediate and propagated influence, controllability distinguishes context that truly shapes the model’s predictions from text that is only superficially relevant or simply proximate in position.

This analysis enables more precise inference-time optimization. Instead of heuristic truncation or embedding-based filtering, controllability-based pruning retains segments that exert substantial control over the state while removing low-influence content. When applied to retrieval-augmented generation, the framework highlights which retrieved passages genuinely guide the model’s reasoning process. It also exposes failure cases where the model relies on stylistically coherent but factually weak sources instead of authoritative material.

\subsubsection{Prompt Engineering and In-Context Learning}
Controllability turns prompt engineering from a trial-and-error exercise into a measurement-driven process. By scoring instructions, demonstrations, and queries, practitioners can quantify which examples meaningfully affect the model’s internal state and which are redundant. This supports systematic selection of demonstrations that maximize influence on the target query and reduces prompt length while preserving task performance. Layer-wise controllability reveals how prompts are processed hierarchically: early layers often capture broad patterns across demonstrations, while deeper layers amplify semantically aligned examples. For Chain-of-Thought (CoT) prompting, comparing controllability with and without reasoning chains identifies which steps genuinely reshape the model’s internal trajectory. This enables refinement of CoT prompts by pruning ineffective steps, elaborating high-impact ones, or reordering reasoning to improve influence on the final prediction.

\subsubsection{Multi-Step Reasoning and Code Generation}
Complex reasoning, such as math problem solving, requires models to maintain and combine information across steps. Controllability analysis provides a temporal map of this process by measuring how each premise or intermediate step influences later conclusions. This exposes reasoning bottlenecks, such as early information that decays too quickly or steps that contribute little to the solution, guiding targeted improvements in model design or prompting strategy. For code generation and program synthesis, controllability traces dependencies across function signatures, type declarations, and prior code blocks. High-influence elements reveal the data flow the model implicitly reconstructs as it generates new code. This insight is useful for debugging: when a model produces incorrect code, controllability maps highlight which parts of the prompt or prior context most strongly drove the erroneous generation, pointing to misinterpreted instructions or inconsistencies in earlier code.

\subsection{Multimodal and Cross-Domain Applications}
The controllability framework can be extended to multimodal and cross-domain SSM applications, offering a unified mathematical tool for interpreting how heterogeneous inputs influence state dynamics. By quantifying how information flows across modalities (vision, language, audio, and temporal signals) the framework reveals how models integrate and prioritize different sources of input. This architecture-agnostic formulation enables consistent interpretability across tasks that require complex reasoning, fusion, or long-range temporal dependencies.

Across modalities such as vision-language modeling, speech and acoustic analysis, scientific forecasting, and biological sequence processing, controllability exposes the underlying mechanisms that drive prediction. It identifies dominant modalities in multimodal fusion, highlights salient temporal regions in audio or time series, and pinpoints biologically or structurally critical subsequences in genomic and protein data. These influence patterns help diagnose failures like incomplete fusion, overreliance on priors, spurious correlations, or loss of long-range dependencies. The framework also applies to reinforcement learning and real-time systems, where controllability reveals which actions most strongly influence future state trajectories and supports interpretable temporal credit assignment. Its computational efficiency, linear in state dimension and sequence length, enables real-time interpretability in production settings without modifying model architecture. By offering a single, scalable method for analyzing information flow across all SSM-based domains, controllability provides a foundation for unified interpretability in next-generation sequence models.

\section{Conclusion and Future Directions}
\label{sec:conclusion}
This work introduces controllability-based analysis as a principled, mathematically rigorous framework for interpreting State Space Models, addressing a critical gap in our understanding of this rapidly emerging architectural paradigm that promises to reshape artificial intelligence across language, vision, time-series, and multimodal domains. By grounding our approach in classical control theory, rather than gradient approximations or attention heuristics, we deliver a fundamentally novel interpretability method that directly quantifies how input elements control the internal state dynamics of SSMs. 

We propose two complementary formulations that balance generality and computational efficiency: (1) a \textit{Jacobian-based approach} that measures influence on aggregated outputs through the full chain of state propagation, applicable to any SSM architecture including those with non-diagonal or structured state matrices, and (2) a \textit{Gramian-based approach} that leverages the analytical solution to the Lyapunov equation for systems with diagonal state matrices, offering substantially faster computation through closed-form evaluation. The Jacobian method provides maximum flexibility and breadth, supporting arbitrary SSM variants, hybrid architectures, and custom state parameterizations, making it the method of choice for exploratory analysis and novel architectures. The Gramian method, while requiring the diagonal assumption satisfied by most modern SSMs (S4, Mamba, and derivatives), achieves superior efficiency by avoiding iterative matrix products, making it ideal for large-scale deployment and real-time interpretability applications.

Through extensive empirical validation on three diverse medical imaging modalities, mammography, dermatology, and hematology, we demonstrate that VSSMs naturally implement hierarchical attention refinement, progressively concentrating influence from diffuse low-level features to semantically meaningful diagnostic patterns. Our analysis reveals domain-specific controllability signatures that align with clinical diagnostic criteria, validates the impact of scanning strategies on spatial attention, and exposes the temporal dynamics of information propagation through state transitions. Beyond medical imaging, we articulate a comprehensive vision for controllability analysis across computer vision (autonomous driving, video understanding, adversarial robustness), natural language processing (long-context modeling, prompt engineering, reasoning), and cross-domain applications, establishing this framework as a foundational interpretability paradigm for the SSM era.

We envision controllability-based interpretability serving as a versatile, high-resolution analytical instrument for understanding the mechanics of modern sequence models. Controllability analysis transforms SSMs from opaque function approximators into transparent dynamical systems whose information flow can be precisely measured, diagnosed, and optimized. This transparency has immediate practical implications for model debugging, safety validation in high-stakes domains, regulatory compliance, and principled architecture design informed by mechanistic understanding rather than empirical trial-and-error. While our work establishes controllability analysis as a powerful and theoretically grounded tool for SSM interpretability, several important limitations and exciting research directions warrant investigation:

\textbf{Non-Linear Dynamics and Inter-Layer Interactions:} The Gramian formulation assumes local linearity of the SSM dynamics, which holds exactly for the linear state transitions $\vec{x}_t = \bar{\mathbf{A}}\vec{x}_{t-1} + \bar{\mathbf{B}}\vec{u}_t$ within each SSM layer. However, modern deep SSM architectures incorporate non-linearities between layers—activation functions (SiLU, GELU), normalization schemes (LayerNorm, RMSNorm), gating mechanisms, and residual connections, that fundamentally alter information flow in ways not captured by per-layer linear analysis. Extending the controllability framework to account for these non-linear transformations represents a significant theoretical challenge. Potential approaches include: (1) differential geometric methods that characterize controllability on manifolds for non-linear dynamical systems, (2) higher-order Taylor expansions around the current state trajectory to approximate local non-linear effects, or (3) empirical Gramian computation via perturbation analysis that directly measures state reachability under non-linear dynamics. Successfully incorporating non-linearities would significantly improve fidelity for complex, deeply stacked SSM architectures.

\textbf{End-to-End Attribution Through Network Depth:} Our current analysis computes controllability independently at each layer, yielding layer-wise influence maps that highlight the hierarchical structure of feature processing. A natural next step is to propagate these influence scores through the entire network to obtain a single, end-to-end attribution map from raw inputs to final outputs. Achieving this requires explicitly modeling information flow across residual connections, pooling operations, and other cross-layer interactions, an inherently non-trivial challenge for SSM-based architectures. While this problem is conceptually related to attention rollout in Transformers, the sequential, state-dependent nature of SSMs introduces additional complexity. Promising directions include adapting techniques such as layer-wise relevance propagation or developing SSM-specific propagation rules that respect state dynamics, potentially enabling a unified global controllability measure.

\textbf{Temporal Causal Analysis and Predictive Explainability:} For sequential data with inherent temporal structure (video streams, sensor time-series, conversational dialogue) the current method provides static, per-time-step influence scores that measure aggregate impact on the final output. A richer analysis would capture \textit{temporal causal influence}: how inputs at time $t$ specifically affect outputs at future times $t+\tau$, enabling causal reasoning and predictive explainability.This requires solving the time-varying Gramian over finite horizons, a computationally more demanding but theoretically tractable problem. It would enable predictive explainability for forecasting tasks, early warning systems, and multi-step reasoning chains where understanding temporal causality is paramount.

\bibliographystyle{unsrt}  
\bibliography{references} 

\end{document}